\documentclass{article}

\pdfoutput=1

\usepackage{graphicx}
\usepackage{bm}
\usepackage{amsmath,amsfonts,amssymb}
\usepackage{epsfig,graphicx,psfrag,color,url,tabularx,booktabs,cite,epstopdf}
\usepackage{tikz,pgf,multirow,diagbox}
\usetikzlibrary{arrows,shapes,snakes,backgrounds,petri}
\usetikzlibrary{positioning}
\usetikzlibrary{calc}

\newcommand{\m}[1]{\boldsymbol{#1}}

\date{15 September 2012}
\begin{document}
\title{Nonparametric Bayesian models of hierarchical structure in
  complex networks}
\author{Mikkel~N.~Schmidt, Tue~Herlau, and Morten~M{\o}rup}
\maketitle

\begin{abstract}
  Analyzing and understanding the structure of complex relational data
  is important in many applications including analysis of the
  connectivity in the human brain.
  Such networks can have prominent patterns on different scales,
  calling for a hierarchically structured model.
We propose two non-parametric Bayesian hierarchical network models
based on Gibbs fragmentation tree priors, and demonstrate their
ability to capture nested patterns in simulated networks.
On real networks we demonstrate detection of hierarchical structure
and show predictive performance on par with the state of the art.
We envision that our methods can be employed in exploratory analysis
of large scale complex networks for example to model human brain
connectivity.
\end{abstract}

\section{Introduction}
Complex networks are an integral part of all natural and man made
systems. Our cells signal to each other, we interact in social
circles, our cities are connected by transportation, water and power
systems and our computers are linked through the Internet. Modeling
and understanding these network structures have become an important
endeavor in order to comprehend and predict the behavior of these many
systems.

A central organizational principle is that entities are
hierarchically\footnote{In this work we define a hierarchy to denote a
  decomposition of a complex relational system into nested sets of
  subsystem rather than as a formal organization of successive sets of
  subordinates~\cite{simon1962}.}  organized and that these
hierarchical structures plays an important role in accounting for the
patterns of connectivity in these
systems~\cite{simon1962,ravasz2003,Roy2007,sales2007,Clauset2008,meunier2010}.
A common notion in modeling complex network is the idea of
\emph{communities}: Groups of nodes that are more densely connected
internally than externally. This notion has led to models of networks
as collections of groups of nodes that determine how edges are
formed~\cite{fortunato2010,MMMNS_BCD}. For instance, people in a
social network may be grouped according to family, workplaces,
schools, or entire countries, and it is assumed these groups determine
the formation of social relations. In many complex systems communities
do not exist only at a single scale but can further be partitioned
into submodules and sub-submodules, i.e., as parts within parts
\cite{simon1962,meunier2010}. For instance, are the communities in a network of
schoolchildren in a city best described by schools or school classes?
How about social cliques within classes or year groups of classes?
It seems that different answers are relevant in different contexts
influencing the scale on which the network should be analyzed; but
discovering the \emph{hierarchical} structure governing such
relationships in a network would tell us more than any particular choice
of resolution.

Previous research on discovering hierarchical structure in networks
has primarily focussed on binary
trees~\cite{Ravasz2002,Ahn2010,Breiger1975,Newman2004,Roy2007,Clauset2008,Roy2009}. Given
a set of nodes and a matrix of affinities between them, a commonly
used tool to uncover their organization is hierarchical clustering
using either agglomerative \cite{Ravasz2002,Ahn2010} or divisive
approaches~\cite{Breiger1975,Newman2004}. These traditional
hierarchical clustering approaches, however, have the following three
major drawbacks~\cite{sales2007}:
\begin{enumerate}
\item They are local in their objective function and do not form a
  well defined global objective.
\item The number of partitions is not well defined and various
  heuristics are commonly invoked to determine this number.
\item The output is always a binary hierarchical tree, regardless of
  the underlying true organization.
\end{enumerate}

Addressing the first two drawbacks, a number of non-parametric
Bayesian models have been proposed: Roy~et~al.~\cite{Roy2007} and
Clauset~et~al.~\cite{Clauset2008} have studied Bayesian generative
models for binary hierarchical structure in networks, assuming a
uniform prior over binary trees, and Roy~and~Teh have proposed the
Mondrian Process~\cite{Roy2009} in which groups are formed by
recursive axis-aligned bisections. Addressing the third drawback,
Herlau~et~al.~\cite{Herlau2012} have proposed using a uniform prior
over multifurcating trees with leafs terminating at groups of network
nodes, and Knowles~and~Ghahramani~\cite{Knowles2011} have mentioned
the applicability of their multifurcating Pitman-Yor Diffusion tree as
a prior for learning structure in relational data.

In this work we propose two non-parametric Bayesian hierarchical
network models based on multifurcating Gibbs fragmentation
trees~\cite{McCullagh2007}. We leverage Bayesian nonparametrics to
devise models that:
\begin{itemize}
\item \emph{Are generative.} This allows us to simulate networks from
  the model, e.g., for use in model checking, and gives a principled
  approach to handling missing data.
\item \emph{Capture structure at multiple scales.} The models
  simultaneously learns about structures from macro scale involving
  the whole network to micro scale involving only a few nodes.
\item \emph{Can infer whether or not hierarchical structure is
    present.} If there is no support for a hierarchy the models
  can reduce to a non-hierarchical structure.
\item \emph{Are consistent and infinitely exchangable.} The models are
  extendable to an infinite sequence of networks of increasing size,
  allowing them to increase and adapt their structure to accomodate
  new data.
\end{itemize}

The paper is structured as follows. In section~\ref{methods} we review
the Gibbs fragmentation tree process~\cite{McCullagh2007} and describe
our models for hierarchical structure in network data. In
section~\ref{results} we analyze the hierarchical structure in
simulated as well as real networks; in particular, we investigate the
support for hierarchical structure in structural whole brain
connectivity networks derived from diffusion spectrum imaging based on
the data provided Hagmann~et~al.~\cite{hagmann2008}. In
section~\ref{conclusion} we present our conclusions and avenues of
further research.

\section{Models and methods}
\label{methods}
\subsection{Fragmentation processes and trees}
Following the presentation in \cite{McCullagh2007}, we review the
multifurcating Gibbs fragmentation tree process. The end result is a
projective family of exchangeable distributions over rooted
multifurcating trees with $n$ leafs.

A rooted multifurcating tree can be represented by a
\emph{fragmentation} of the set of leafs. Let $B$ be the set of leafs
and $n = |B|$ the total number of leafs. Recall that a
\emph{partition} $\pi_B$ of $B$ is a set of $2$ or more non-empty
disjoint subsets of $B$, $\pi_B=\{B_1, B_2, \dots, B_k\}$, such that
the union is $B$. In the following we denote the size of these subsets
by $n_i=|B_i|$. A \emph{fragmentation} $T_B$ of a set $B$ is a
collection of non-empty subsets of $B$ defined recursively such that
the set of all nodes is a member, $B\in T_B$; each member of the
partition $\pi_B$ is a member, $B_1\in T_B$, $\dots$, $B_k\in T_B$;
each member of partitions of these subsets are members, and so on
until we reach the singletons. Recursively we may
write~\cite{McCullagh2007}
\begin{equation}
  T_B=\left\{
    \begin{array}{ll}
      \{B\},
      & |B|=1,\\
      \{B\} \cup T_{B_1} \cup \dots \cup T_{B_k},
      & |B|\ge 2.
    \end{array}
  \right.
\end{equation}
For example, the tree in Figure~\ref{fig:InfiniteTree} which has leafs
$B=\{1,2,3\}$ is represented by the fragmentation
$T_B=\big\{\{1,2,3\},\{1,2\},\{1\},\{2\},\{3\}\big\}$. Uniquely
associated with the fragmentation is a multifurcating tree where each
element in $T_B$ above serves as a node: $B$ is the root node, and the
singletons are the leafs.  To emphasize this connection, $T_B$ is
called a \emph{fragmentation tree}~\cite{McCullagh2007}.  The
collection of all fragmentation trees for a set $B$ is denoted by
$\mathbb{T}_B$.

Let $A \subset B$ be a nonempty proper subset of the leaf
nodes. The \emph{restriction} of $T_B$ to $A$ is defined as ``the
fragmentation tree whose root is $A$, whose leaves are the singleton
subsets of $A$ and whose tree structure is defined by restriction of
$T_B$.''~\cite{McCullagh2007}. This is also called the
\emph{projection} of $T_B$ onto $A$ and denoted by $T_{B,A}$.

A \emph{random fragmentation model}~\cite{McCullagh2007} assigns a
probability to each tree $T_B\in\mathbb{T}_B$ for each finite subset
$B$ of $\mathbb{N}$. The model is said to be:
\begin{itemize}
\item \emph{Exchangeable} if the distribution of $T_B$ is invariant to
  permutations on $B$, i.e., the distribution does not depend on the
  labelling of the leaf nodes.
\item \emph{Markovian} if, for a given $\pi_B=\{B_1,B_2,\dots,B_k\}$,
  each of the $k$ restricted trees $T_{B_i,B}$ are independently
  distributed as $T_{B_i}$.
\item \emph{Consistent} if, for all nonempty $A \subset B$, the
  projection of $T_B$ onto $A$ is distributed like $T_A$.
\end{itemize}

The starting point for constructing a random fragmentation model is a distribution over partitions of $B$.
By exchangeability this
distribution must be a symmetric function depending only on the size
of each subset,
\begin{align}
  q(\pi_B) = q\big(n_1,\dots,n_k\big),
\end{align}
where $q$ is called the \emph{splitting rule}. Abusing notation, we
write the splitting rule as a function of a partition or equivalently
as a function of the sizes of the subsets in the partition.

\begin{figure}
  \centering
  \includegraphics[width=\columnwidth]{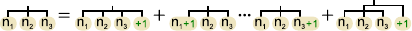}
  \caption{A Markovian consistent splitting rule satisfies the
    condition that the probability of a partition is equal to the sum
    of the probabilities of all configurations where a single extra
    node is added.}
  \label{fig:consistent}
\end{figure}

Requiring Markovian consistency places a further constraint on the
splitting rule~\cite{McCullagh2007} (see Figure~\ref{fig:consistent}),
\begin{multline}
  q\big(n_1,\dots,n_k\big) =  q\big(n_1,\dots,n_k,1\big) +\\
  q\big(n_1+1,\dots,n_k\big) + \cdots + q\big(n_1,\dots,n_k+1\big) +\\
  q\big(1,n_1+\cdots+n_k\big)q\big(n_1,\dots,n_k\big).
\end{multline}

McCullagh~et~al.~\cite{McCullagh2007} show that under the further
condition that the splitting rule is of Gibbs form,
\begin{equation}
  q\big(n_1,\dots,n_k\big) = \frac{a(k)}{c\big(n\big)}\prod_{i=1}^{k}w(n_i),
\end{equation}
where $w(\cdot)\ge 0$ and $a(\cdot)\ge 0$ are some sequences of
weights and $c(\cdot)$ is a normalization constant. Specifically the only
admissible splitting rule is given by\footnote{McCullagh~et~al.~\cite{McCullagh2007} do not give an explicit formula for the normalization constant.}
\begin{align}
  q(n_1,\dots,n_k) & = \frac{(\tfrac{\beta}{\alpha})^{(k)}(-1)^k}
  {\beta^{(n)}-\tfrac{\beta}{\alpha}(-\alpha)^{(n)}}\prod_{i=1}^k(-\alpha)^{(n_i)},
  \label{eq:SplittingRule}
\end{align}
where $x^{(y)}=x(x+1)\cdots(x+y-1)=\tfrac{\Gamma(x+y)}{\Gamma(x)}$
denotes the rising factorial. The splitting rule has two parameters,
$\alpha\ge 0$ and $\beta \ge -\alpha$. To simplify the notation in the
following we define $q(0)=q(1)=1$.

By the Markovian property the distribution over fragmentations can
then be characterized as a recursive product of these splitting rules,
one for each set in the fragmentation or equivalently for each node in
the tree. This gives rise to the following representation of all
exchangeable, Markovian, consistent, Gibbs fragmentation processes,
\begin{align}
  p(T_B) & = \prod_{A \in T_B} q(\pi_A),
  \label{eq:gibbsfrag}
\end{align}
where $q(\cdot)$ is given by Eq.~(\ref{eq:SplittingRule}) and $\pi_A$
denotes the children of node $A$ in $T_B$.

\begin{figure}[t!]
  \centering
  \includegraphics[width=\columnwidth]{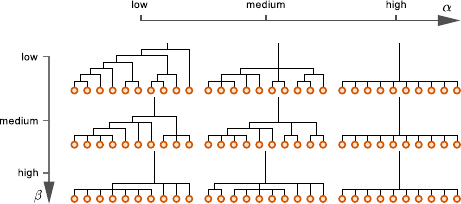}
  \caption{Multifurcating trees generated from the two-parameter Gibbs
    fragmentation tree process. The parameters govern the distribution
    of the degree of the internal nodes in the tree. The trees shown
    correspond to $\alpha\in\{0.1, 0.5, 1\}$ and
    $\beta+\alpha\in\{0.1, 1, 10\}$.}
  \label{fig:prior}
\end{figure}

To illustrate how the properties of the Gibbs fragmentation tree
distribution is governed by the two parameters $\alpha$ and $\beta$ we
have generated a few trees from the distribution, varying the
parameters within their range (see Figure~\ref{fig:prior}).

\subsection{Relation to the nested CRP}
The Gibbs fragmentation tree is closely related to the two-parameter
version of the Chinese restaurant process (CRP)~\cite{aldous1985}. The
CRP is a partition-valued discrete stochastic process: For $n=1$
element, the CRP assigns probability one to the trivial partition. As
$n$ increases, element number $n+1$ is added to an existing set of elements in the partition with
probability $\frac{n_i-\alpha}{n+\beta}$ or added to the partition as a new singleton set
with probability $\frac{\beta+k\alpha}{n+\beta}$. Taking the
product of $n$ such terms yields the expression for the probability
assigned to a given partition $\pi_B$,
\begin{equation}
  p(\pi_B) = \frac{\Gamma(\beta)\alpha^k\Gamma\big(\tfrac{\beta}{\alpha}+k\big)}
  {\Gamma(\beta+n)\Gamma\big(\tfrac{\beta}{\alpha}\big)}
  \prod_{i=1}^{k}\frac{\Gamma(n_i-\alpha)}{\Gamma(1-\alpha)}.
  \label{eq:crp}
\end{equation}
As the number of elements goes to infinity, the CRP defines a
distribution over partitions of a countably infinite set.

\begin{figure}
  \centering
  \includegraphics[width=60mm]{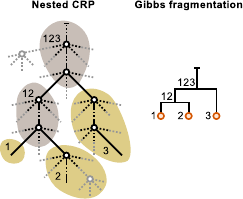}
  \caption{Illustration of the relation between the nested Chinese
    restaurant process (CRP) and its finite representation as a Gibbs
    fragmentation tree. In the nested CRP, each internal node in the
    tree splits into an infinite number of subtrees. Each element
    associated with the tree traces a infinite path starting at the
    root. In the illustration three elements are associated with the
    tree; thus, the hierarchical structure relating the observables
    (which is what we are ultimately interested in learning) can be
    represented by a Gibbs fragmentation tree of finite size. As an
    example, in the finite representation the root node (labeled
    ``123'') corresponds to the first common ancestral node of all
    observables as well as the parents and grandparents etc. of that
    node all the way to the root of the tree.}
  \label{fig:InfiniteTree}
\end{figure}

Now, consider a set of nested Chinese restaurant processes as proposed
by Blei~et~al.~\cite{Blei2007,Blei2004}: First, the set $B$ is
partitioned into $\pi_B =\{B_1,\dots,B_k\}$ according to a CRP. Next,
each subset $B_i$ is partitioned again according to a CRP with the
same parameters, and the process is continued recursively ad infinitum
(see Figure~\ref{fig:InfiniteTree}). This nested CRP thus defines ``a
probability distribution on infinitely deep, infinitely branching
trees.''~\cite{Blei2007}. Blei~et~al.\ use this nested CRP as a prior
distribution in a Bayesian non-parametric model of document
collections by assigning parameters to each node in the tree and
associating documents with paths through the tree.

In the nested CRP, each element traces an infinite path through the
tree. When a finite number of elements $n$ is considered, they trace a
tree of finite width but infinite depth. In the terminology of the
random fragmentation model, the nested CRP model corresponds to a
fragmentation tree using a CRP as a splitting rule. The key difference
between the Gibbs fragmentation trees and the nested CRP is that the
CRP splitting rule allows fragmenting into the trivial partition,
i.e., it allows nodes with a single child whereas the Gibbs fragmentation tree allways has at least two children. Instead of
working directly with this infinitely deep tree, we can consider the
equivalence class of trees with the same branching structure by
marginalizing over the internal nodes that do not branch out, yielding
a tree of finite depth. The distribution for this equivalence class
can be arrived at by marginalizing over the number of consecutive
trivial partitions that occurs before the first ``real'' split. According
to the CRP in Eq.~(\ref{eq:crp}), the trivial partition has
probability
\begin{equation}
  p_0 \equiv p\left(\pi_B=\{B\}\right) =
  \frac{\Gamma(\beta)\alpha\Gamma\big(\tfrac{\beta}{\alpha}+1\big)}
  {\Gamma(\beta+n)\Gamma\big(\tfrac{\beta}{\alpha}\big)}
  \frac{\Gamma(n-\alpha)}{\Gamma(1-\alpha)}.
  \label{eq:trivialcrp}
\end{equation}
We wish to marginalize over seeing zero, one, two, etc.\ trivial
partitions before the first split. To compute this marginalization,
the CRP distribution must be multiplied by
\begin{multline}
  1+p_0+p_0^2+\dots = \frac{1}{1-p_0}
  =\frac{\beta^{(n)}}{\beta^{(n)}+\tfrac{\beta}{\alpha}(-\alpha)^{(n)}},
  \label{eq:geometric}
\end{multline}
where we have inserted Eq.~(\ref{eq:trivialcrp}) and used the
geometric series formula to compute the infinite sum. Multiplying
Eq.~(\ref{eq:geometric}) by the CRP in Eq.~(\ref{eq:crp}) yields
exactly the splitting rule of McCullagh~et~al.\ in
Eq.~(\ref{eq:SplittingRule}) establishing the relation between the
Gibbs fragmentation tree and the nested CRP.

\subsection{Tree-structured network models}
We now turn to applying the Gibbs fragmentation process as a prior in
a Bayesian model of hierarchical structure in complex networks.  To
simplify the presentation we focus on simple graphs but note that the
main ideas can be extended to more intricate relational data such as
weighted and directed graphs etc. A simple graph with $n$ nodes can be
represented by a symmetric binary adjacency matrix $\m A$ with element
$a_{i,j}=1$ if there is a link between node $i$ and $j$.

First, consider a model in which each possible link $a_{i,j}$
is generated independently from a Bernoulli distribution (a biased
coin flip) with probability $\theta_{i,j}$. Since each possible link
has its own parameter, no information is shared in the model between
different nodes and links and the model will not be able to
generalize. Combining information between network nodes is necessary,
for example by pooling the parameters for blocks of similar nodes. The
particular way in which these parameters are shared is the key
difference between the models we discuss here. In the stochastic
blockmodel~\cite{snijders1997,holland1983} network nodes are clustered
into blocks which share their probabilities of linking within and
between blocks. The infinite relational model
(IRM)~\cite{Kemp06,xu2006} is a nonparametric Bayesian extension of
the stochastic blockmodel based on a CRP distribution over block
structures. We consider the IRM model the state of the art in
Bayesian modeling of large scale complex networks.

The link probabilities \emph{between} the blocks in these models can
either be individual for each pair of blocks (unpooled) as in the IRM
model or be completely shared as a single between-block link
probability (complete pooling) as in \cite{Hofman2008}. Furthermore
the model can specify that blocks have more internal than external
links leading to an interpretation of the blocks as communities of
highly interconnected nodes~\cite{morupschmidt2012}.

\begin{figure}
  \centering
  \includegraphics[width=\columnwidth]{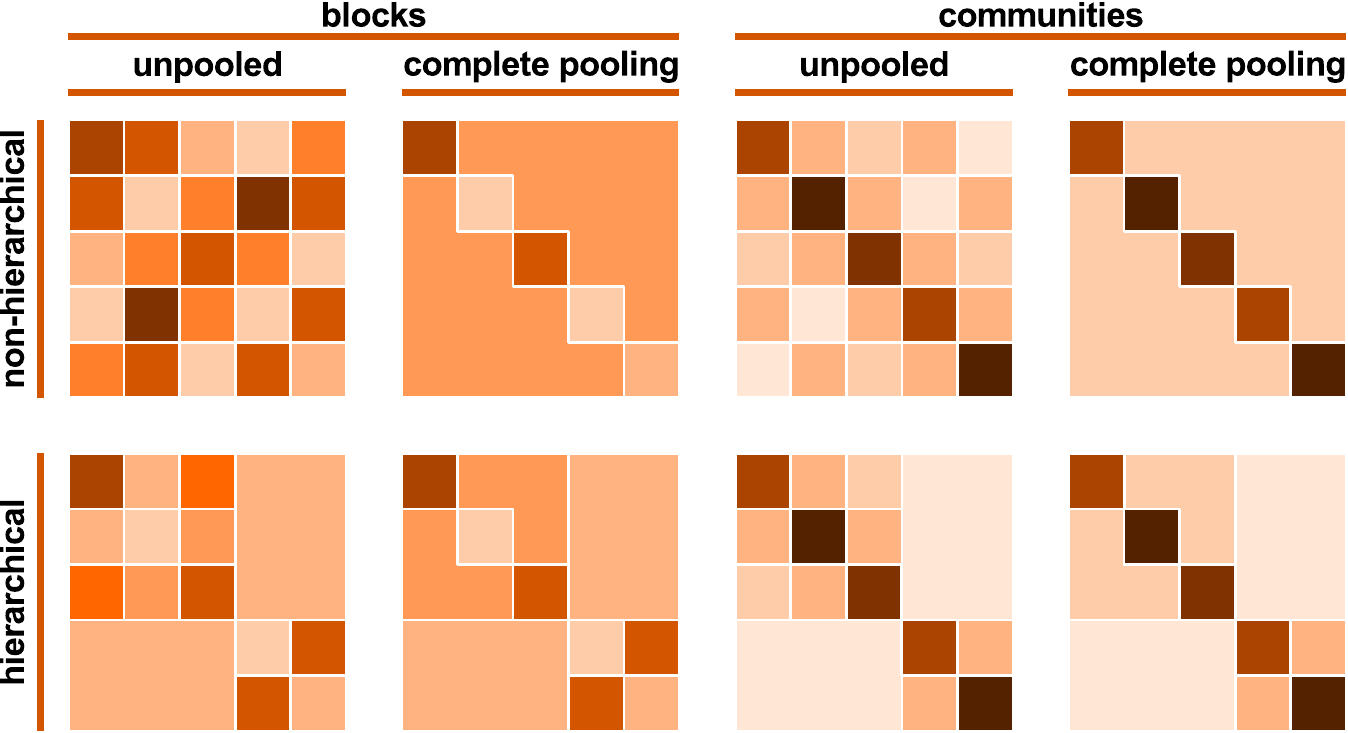}
  \caption{Illustration of different approaches to modelling
    (hierarchical) group structure in complex networks. The figures
    shows matrices of probabilities of links between groups in a
    network with five groups (darker color indicates higher link
    probability). Groups of nodes can be allowed to link to other
    groups with independent probabilities (denoted blocks), or
    restricted to have higher probability of links within than between
    groups (denoted communities). Furthermore, between-group link
    probabilities can be independent (unpooled) or shared (pooled)
    amongst all groups at each level of the hierarchy.}
  \label{fig:blocks}
\end{figure}

\begin{figure}
  \centering
  \includegraphics[width=\columnwidth]{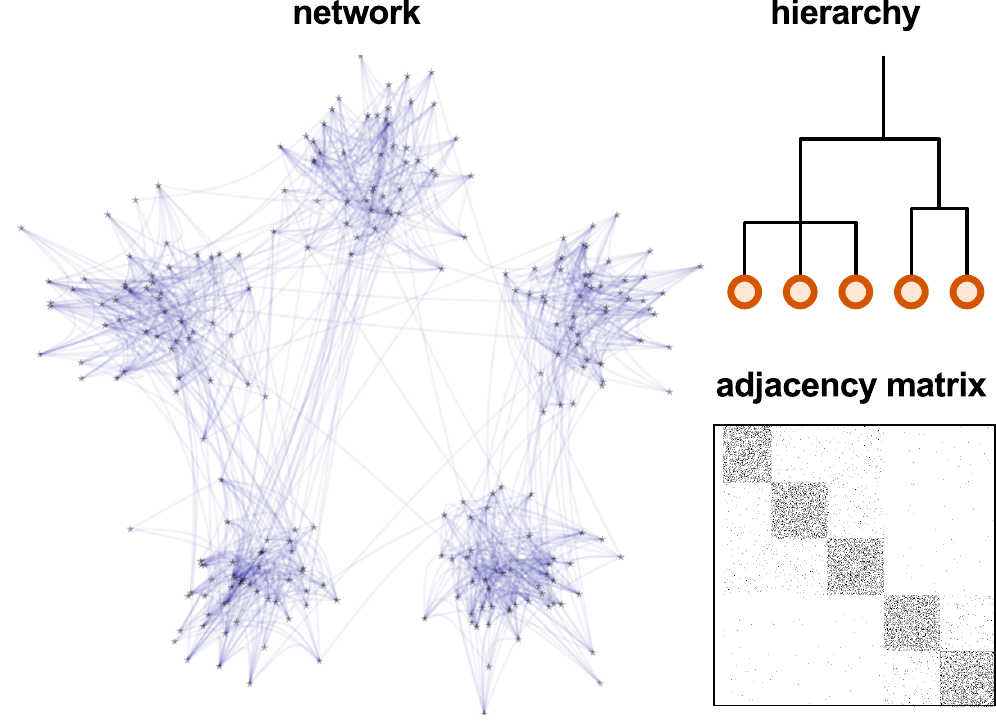}
  \caption{Simulated example of a complex network with hierarchical
    group structure. The network has five clusters; however, three of
    these (top and left) are more connected to each other than to the
    remaining two clusters, forming a super cluster. The goal of this
    work is to automatically detect such hierarchical structure and
    learn from data the number of clusters, their hierarchical
    organization, as well as the depth of the hierarchy.}
  \label{fig:graph}
\end{figure}

The hierarchically structured models of complex networks proposed
here correspond to nested stochastic blockmodels in which each block
is recursively modelled by a stochastic blockmodel, and we use the
Gibbs fragmentation tree process as a prior over the nested block
structure. As in the stochastic blockmodel, links between blocks can
be pooled or not, leading to models with different
characteristics. Figure~\ref{fig:blocks} illustrates these different
approaches to pooling parameters in block structured network models,
and Figure~\ref{fig:graph} illustrates a network that can be well
characterized by a hierarchical block structure.

Let $\m A$ denote the observed network and let $T$ denote a
fragmentation of the network nodes. The following general outline of a
probabilistic generative process can be used to characterize a complex
network with a hierarchical cluster structure.
\begin{enumerate}
\item Generate a rooted tree $T$ where the leaf nodes corresponds to
  the vertices in the complex network,
  \begin{equation}
    T\sim p(T|\tau).
  \end{equation}
  Each internal node in the tree corresponds to a cluster of network
  vertices.
\item For each internal node in the tree, generate parameters $\m\theta$
  that govern the probabilities of edges between vertices in each of
  its children,
  \begin{equation}
    \m\theta\sim p(\m\theta|T,\rho).
  \end{equation}
\item For each pair of vertices in the network, generate an edge with
  probability governed by the parameters located at their common
  ancestral node in the tree,
  \begin{equation}
    \m A\sim p(\m A|\m\theta,T,\xi).
  \end{equation}
\end{enumerate}
Several existing hierarchical network models
\cite{Clauset2008,Herlau2012,Roy2007} are special cases of this
approach with different choices for the distributions of $T$, $\theta$, and
$\m A$. Inference in these models entails computing the posterior distribution
over the latent tree structure,
\begin{align}
  p(T|A) = \int \! \frac{p(\m A | \m \theta, T)p(\m \theta |
    T)p(T)}{p(\m A)} d\m \theta .
  \label{eqn:basicbayes}
\end{align}

In the following, we consider two hierarchical block models: An
unpooled and a pooled model (see Figure~\ref{fig:blocks}). As a
distribution over trees we use the Gibbs fragmentation model in
Eq.~(\ref{eq:gibbsfrag}). The likelihood for both models can be
written as a product over all internal nodes in the tree,
\begin{align}
  p(\m A | \m \theta, T) = \prod_{\substack{B \in T\\|B| \geq 2}}
  f_B(\m A, \m \theta_{B},B).
  \label{eq:likelihood}
\end{align}
Assume that the set $B$ according to $T$ fragments into the partition
$\{B_1,B_2,\dots,B_k\}$ and $\ell$, $m$ denote indices of each
fragment. The likelihood then has the following form,
\begin{equation}
  f_B(\m A, \m \theta_{B},B) =
  \prod_{\substack{B_\ell\in B\\B_m\in B\\l<m}}
  \prod_{\substack{i\in B_\ell\\j\in B_m\\i<j}}
  \mathrm{Bernoulli}(\theta_{B,\ell,m}),
\end{equation}
where the products go over each possible link between each possible
pair of blocks.
In the pooled model $\theta_{B,\ell,m} \equiv \theta_B$ are equal for
all blocks, and in the unpooled models, $\theta_{B,\ell,m}$ are
independent. We use independent Beta priors for the link probabilities,
\begin{equation}
  p(\theta_{B,\ell,m}) = \mathrm{Beta}(\rho^+, \rho^-).
\end{equation}
The hyperparameters in our model are $\tau=\{\alpha,\beta\}$ and
$\rho=\{\rho^+,\rho^-\}$. In all experiments these were fixed at
$\alpha=\beta=\tfrac{1}{2}$ and $\rho^+=\rho^-=1$.

\subsection{Implementation}
Inference in the models is performed using Markov chain Monte Carlo
sampling. Due to the conjugacy between the prior and likelihood for
the link probabilities $\theta$, they can be analytically marginalized
allowing collapsed sampling of the tree.

We use the Metropolis-Hastings algorithm with subtree pruning and regrafting (SPR) proposals in which a subtree is removed from the tree and inserted in a new position.
Assume $k$ is a node removed from a tree $T$. Let $T_k$ be the corresponding subtree rooted at $k$ and $T_{\setminus k}$ the tree obtained by projecting out $B_k$. It is useful to distinguish between two types of insertion operations acting on a node $h$ in $T_{\setminus k}$: In moves of type 1 the tree is modified by simply adding $T_k$ as a child to node $h$. Notice that this require $h$ to have at least two children, consequently, for the chain to be ergodic we must include moves of type 2 where
 $T_{\setminus k}$ is modified by replacing the subtree of $T_{\setminus k}$ rooted at $h$, $T_{\setminus k,h}$, with a new subtree with $T_{\setminus k,h}$ and $T_{k}$ as its only children.

While one can simply select between all available insertion operations at random, the hierarchical organization of the network implies that it is rarely prudent to propose moves which move nodes far from where they are attached. We propose an alternative scheme where nodes are removed at random, but the set of allowed insertion moves is selected by taking the parent of the removed node, collecting all vertices in the reduced tree with a travel-distance less than or equal to two from the parent and forming the set of insertion moves of type 1 or 2 as they apply. In all simulations we choose between the two types of proposal moves with probability $\frac{1}{2}$.
\begin{figure}
  \centering
  \begin{tikzpicture}
  \node[inner sep=0] at (0cm,0cm) {\includegraphics[width=7cm]{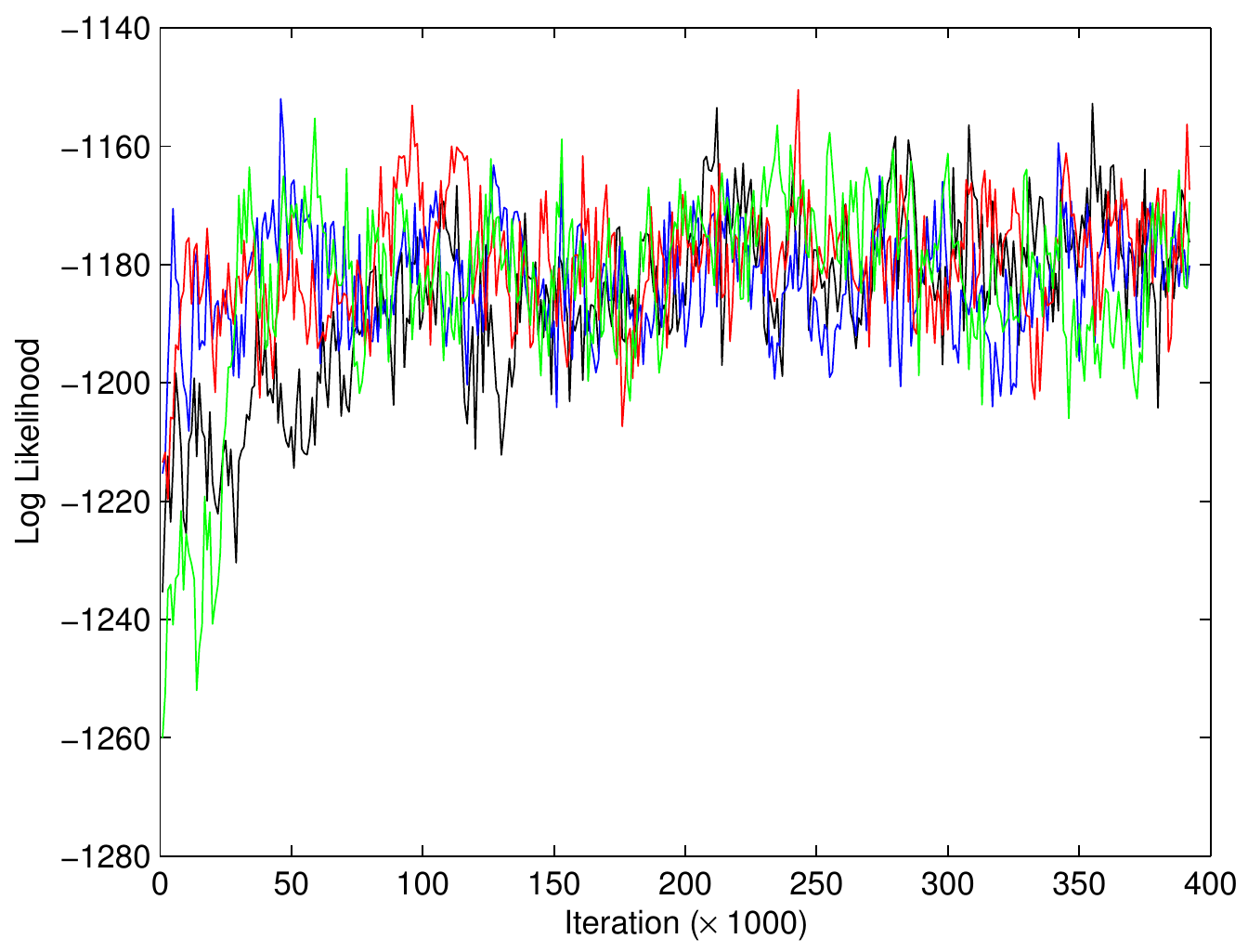}};
  \node[inner sep=0] at (1cm,-0.9cm) {\includegraphics[width=4cm]{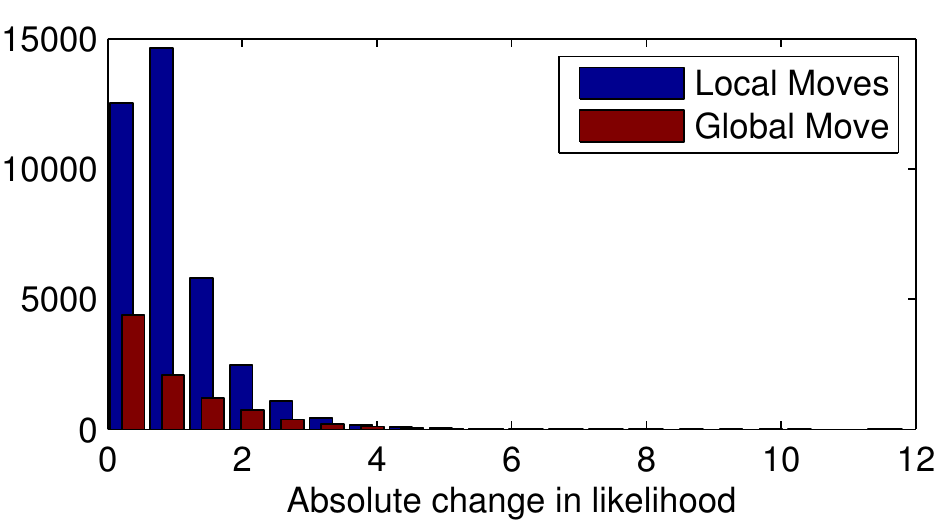}};
  \end{tikzpicture}
  \caption{Log likelihood for 5 chains (shown in different color) running scan 1 of the diffusion datasets of figure \ref{fig:subject1}. The frequent crossing of chains indicate reasonable mixing. Insert show histogram of (absolute) change in log likelighood for accepted moves of the two move classes, blue bars are local moves and red global. Moves to similar state not included.}\label{fig:convergence}
 \end{figure}

By caching counts of links and non-links within each block throughout the tree,  computing only the relevant updates of these counts as a move is accepted, both the pooled and unpooled model can be implemented efficiently. Removing or adding a subtree only changes cached terms associated with the path from the removed/added node to the root. While the unpooled model in the worst case contains $\frac{1}{2}n(n-1)$ $\theta$-parameters (corresponding to the tree where the root immediately splits into $n$ leafs), each is only associated with a single variable (the presence or absence of the single associated link) and their total contribution to the likelihood is linear in the total number of links and non-links and so one only need to keep track of these two values. In general, a large number of singleton clusters is not a computational problem, as only the sum of links and non-links between these clusters must be computed and updated.

In all experiments described in the following section, the sampler was
run for 400'000 iterations. The first half of the samples were
discarded for burn in, and the second half were subsampled by a facter
of 1000, yielding 200 posterior samples. Figure~\ref{fig:convergence}
shows the result of running 5 different chains on the human brain
connectivity dataset described later. As can be seen the likelihood of
each chain frequently change value suggesting that the chains mix. The
insert in the figure shows a histogram of the change in absolute value
of log likelihood for the global and local proposal moves. The local
moves are accepted more than twice as often as the global ($0.093$
compared to $0.036$) and also contribute to significant change in log
likelihood. These results were similar for all the conducted
experiments.
\section{Results}\label{results}
\subsection{Simulations}
\begin{figure*}
\centering
\includegraphics[width=\columnwidth]{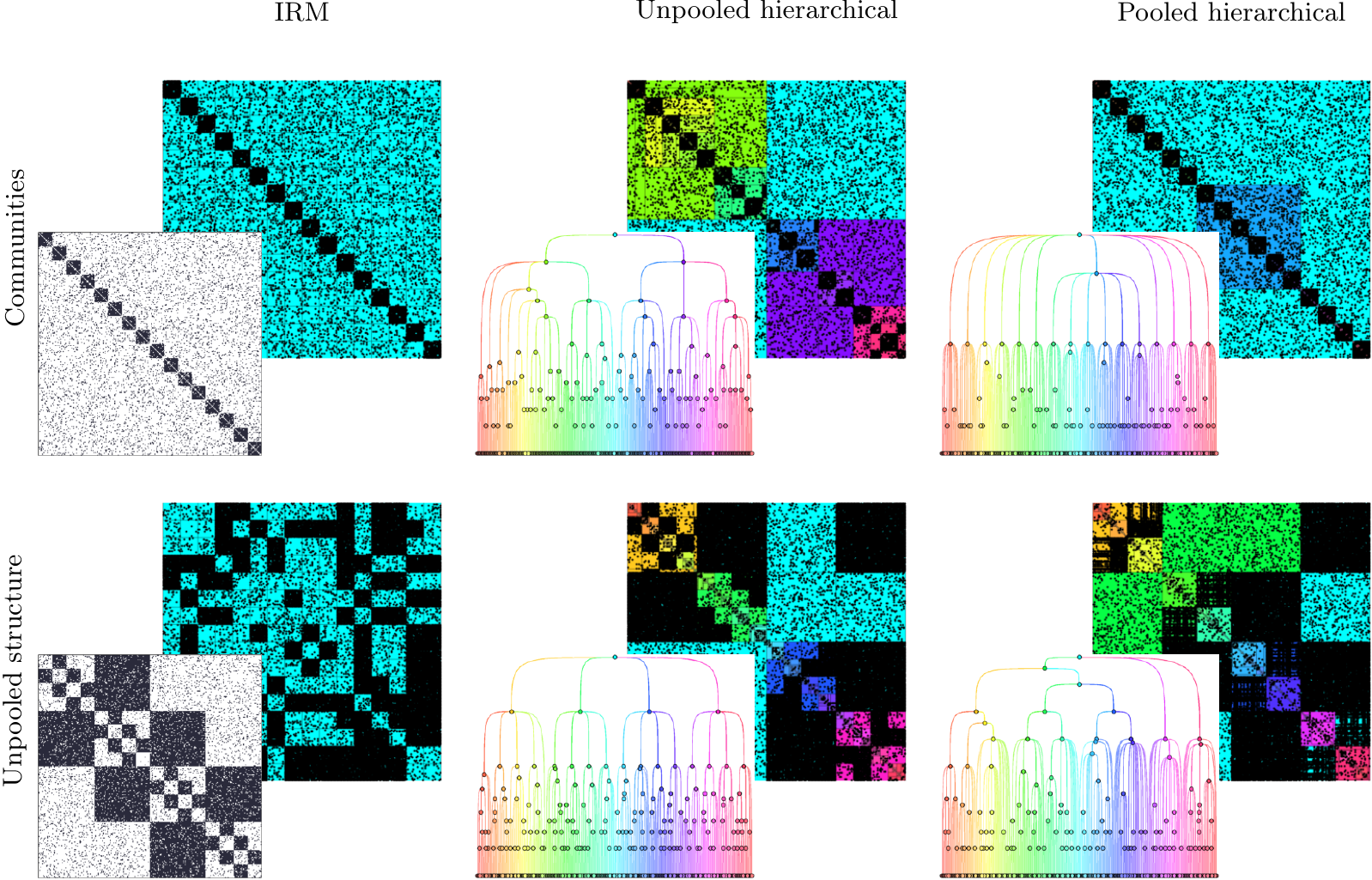}
\caption{IRM, unpooled and pooled hierarchical modeling of four
  synthetic networks. The first network is generated to have strict
  community structure whereas the second network is generated
  according to the IRM model. The third and fourth networks are
  generated according to the unpooled and pooled hierarchical models
  respectively. Colors on the vertices of the inserted tree and on the corresponding adjacency matrix indicate where in the tree the parameters which explain the links reside.}
\label{fig:adat}
\end{figure*}

The proposed models were evaluated and compared with the IRM model on four artificial networks each
chosen with approximately the same overall difficulty but supporting
different types of structure, see also Figure~\ref{fig:adat}. The
first synthetic network (\emph{Communities}) is a network of strict
community structure which can be well accounted for by all models. The
second network (\emph{IRM}) was generated according to the IRM model,
the third (\emph{Unpooled}) according to the unpooled hierarchical
model and the fourth (\emph{Pooled}) according to the pooled
hierarchical model.

In Figure~\ref{fig:adat} it can be seen that the three models well infer structure in the networks they are designed for. We further see that the unpooled model is able to account for the structure of both the IRM model and the pooled model as it is closely related to the IRM model when forming a flat hierarchy while being more flexible than the pooled model when inferring hierarchical structure and is thereby able to account for the pooled hierarchical structures. Consequently, the unpooled model is able to infer the presence of hierarchical structure while reducing to a flat hierarchy corresponding to the IRM model when no such structure is supported by the data. The pooled model has a substantially reduced parameter space compared to the unpooled model. It is better able to identify structure when data indeed supports this type of hierarchical structure while it creates spurious results when the assumption of a pooled hierarchy fails: In the IRM and Unpooled networks, the pooled model clearly underfits.

In Table~\ref{tab:adat} we inspect the models ability to predict
structure in multiple networks generated according to the true model
parameters used to generate the networks in Figure~\ref{fig:adat}. We
trained each models on a single synthetic network of each type and
used the trained model to predict 10 other synthetic networks
generated from the same distribution.  We evaluate the predictive
performance in terms of the mean predictive log-likelihood and average
area under curve (AUC) of the receiver operator characteristic (ROC)
across the samples using link prediction on the complete network. The
models' predictive performance are essentially equal except for the
two cases where the pooled hierarchical model underfits.

\begin{table*}
  \centering
\rotatebox{90}{
  \begin{tabular}{lcccccccc}
    \toprule
    \multirow{2}{*}{\diagbox{Model}{Network}}
             & \multicolumn{2}{c}{Communities} & \multicolumn{2}{c}{IRM}     & \multicolumn{2}{c}{Unpooled} & \multicolumn{2}{c}{Pooled}                                                         \\
             & $\log{L}$                         & AUC                           & $\log{L}$                      & AUC              & $\log{L}$    & AUC              & $\log{L}$    & AUC              \\
    \midrule
    \multirow{2}{*}{IRM}      & -10710       & 0.6769           & -10700       & 0.8964           & -10720       & 0.8989           & -10680       & 0.826            \\
                              & $\pm$ 40       & $\pm$ 0.00138      & $\pm$ 36.7     & $\pm$ 0.000572     & $\pm$ 42.8     & $\pm$ 0.000751     & $\pm$ 41.7     & $\pm$ 0.000867     \\
    \midrule
    \multirow{2}{*}{Unpooled} & -10810 & 0.675    & -11010 & 0.8947  & -10780 & 0.8995 & -10790   & 0.8258 \\
                              & $\pm$ 19.4 & $\pm$ 0.0008    & $\pm$ 49.5 & $\pm$ 0.00077  & $\pm$ 26.1 & $\pm$ 0.000435 & $\pm$ 28   & $\pm$ 0.000684 \\
    \midrule
    \multirow{2}{*}{Pooled}   & -10710 & 0.6763 & -16440 & 0.8155 & -15920 & 0.8207 & -10700 & 0.8256 \\
                              & $\pm$ 23.4 & $\pm$ 0.000664 & $\pm$ 23.4 & $\pm$ 0.000491 & $\pm$ 42.6 & $\pm$ 0.000889 & $\pm$ 25.5 & $\pm$ 0.000587 \\
    \bottomrule
    \end{tabular}}
    \caption{Average predictive log likelihood ($\log{L}$) and AUC
      scores of the three models performance on the four types of
      networks generated. The models are trained on the networks
      generated in figure~\ref{fig:adat} and evaluated on 10
      additional networks generated according to the true model
      parameters used to generate the data given in the figure. }
    \label{tab:adat}
\end{table*}

\subsection{Modeling hierarchical structure in real world networks}
In order to qualitatively evaluate the proposed hierarchical models' ability to account for structure in real networks we consider the following four networks
\begin{itemize}
\item \emph{NIPS:} The NIPS network is a binary graph with a total of
  598 undirected links between the top 234 collaborating Neural
  Information Processing Systems (NIPS) authors in NIPS volumes 1 to
  17 (also analyzed in \cite{miller2009,morupschmidt2012}).
\item \emph{Football:} Network of American football games
  between Division IA colleges during regular season Fall 2000. The
  network consist of 115 colleges and 613 games \cite{girvan2002}.
\item \emph{Les Miserable:} Network of the 254 co-appearances of 77
  characters in the novel Les Miserables \cite{knuth1993stanford}.
\item \emph{Zachary:} Social network of 78 recorded friendships
  between 34 members of a karate club at a US university in the 1970s
  \cite{zachary1977}.
\end{itemize}

\begin{figure*}
  \centering
  \includegraphics[width=\columnwidth]{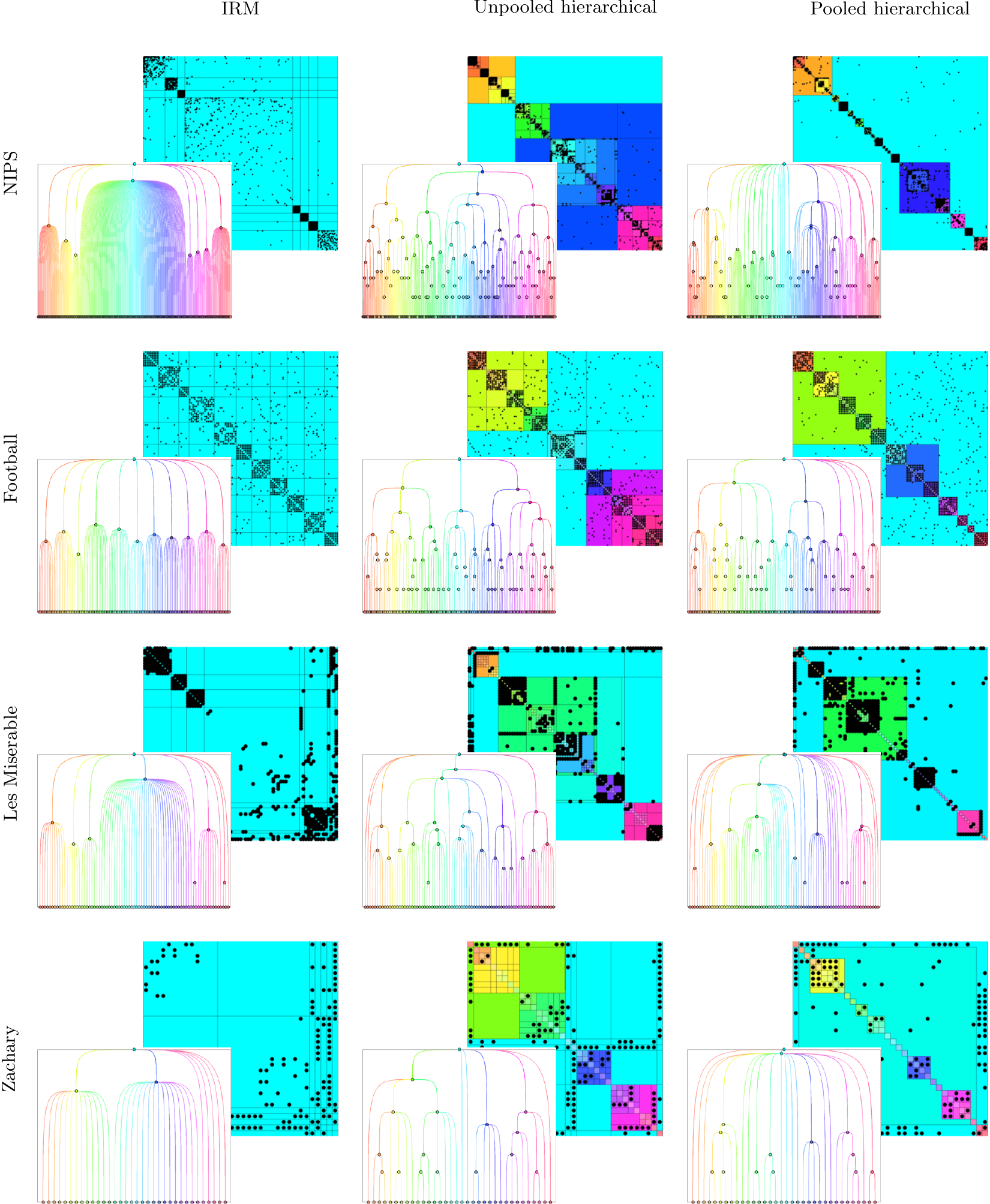}
  \caption{Analysis of the NIPS, Football, Les Miserable and Zachary
    networks by the IRM model as well as the unpooled and pooled
    hierarchical models.}
\label{fig:realdag}
\end{figure*}

The results of the modeling using the IRM model as well as the unpooled and pooled hierarchical models are given in Figure~\ref{fig:realdag}. All models appear to identify network homogeneities. In the NIPS and Les Miserable networks the IRM model extracts a large cluster that group nodes that are not well connected to each other into what more or less appear to represent a noise cluster. This has previously been reported in \cite{ishiguro2012} where it was proposed to extend the IRM model to explicitly account for these clusters representing noise. Rather than treating these nodes as coming from a noise cluster, both the unpooled and pooled hierarchical models are able to detect structure at a level of resolution that is substantially smaller than what the IRM model accounts for and terminate at a level where these nodes in fact form small groups of tightly connected communities. In addition both the pooled and unpooled models are able to represent structure in the graphs in terms of hierarchies and it is observed that many of the splits are multifurcating having three or more children. This well supports the notion that hierarchical structure go beyond the strict binary hierarchies considered in \cite{Clauset2008,Roy2007,Roy2009}. This has also been observed previously when modeling feature data by multifurcating hierarchies \cite{BluTehHel2011a}. Thus, both the pooled and unpooled models identify prominent multifurcating hierarchical structures in the considered networks. These results support the existing literature arguing that many real world networks exhibit hierarchical structure \cite{simon1962,ravasz2003,Roy2007,sales2007,Clauset2008,meunier2010}.

\subsection{Testing for hierarchical structure in whole brain structural connectivity}
Brain networks are believed to exhibit hierarchical modularity, i.e. modules of the brain do not exist only at a single organizational scale but each module are further partitioned into submodules~\cite{meunier2010}. This type of hierarchical organization has been demonstrated to occur in both functional~\cite{meunier2009,ferrarini2009} and structural brain networks~\cite{bassett2010}. We presently investigate if our models indeed support the notion of hierarchical modularity in data of structural brain connectivity.

\begin{figure*}
  \centering
  \includegraphics[width=\columnwidth]{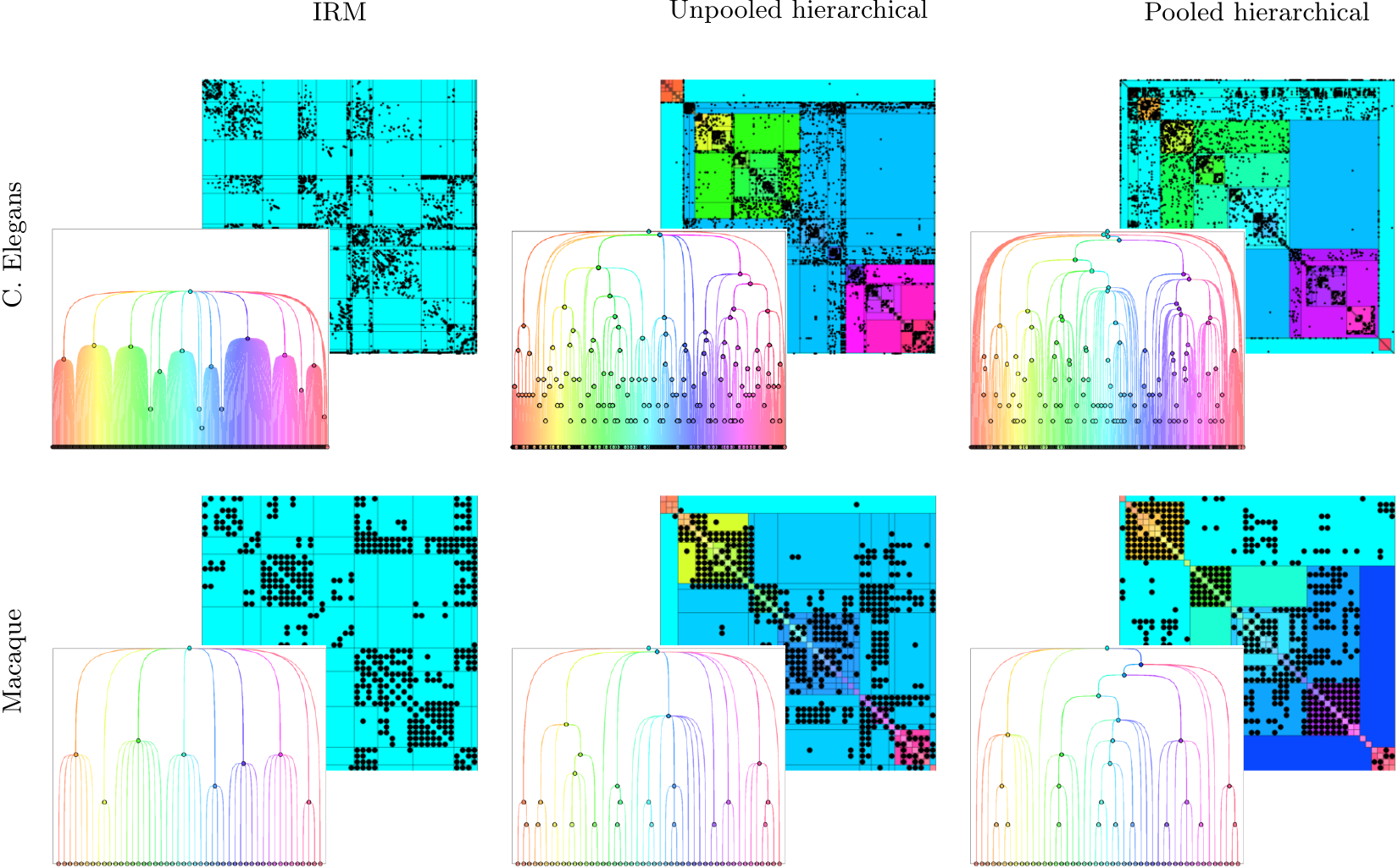}
  \caption{Analysis of the full connectome of
    C. Elegans~\cite{achacoso1992,watts1998} as well as the Macaque
    monkey right hemisphere~\cite{hagmann2008} by the IRM model as
    well as the unpooled and pooled hierarchical models.}
  \label{fig:CELEGANSandMonkey}
\end{figure*}

In Figure~\ref{fig:CELEGANSandMonkey} we analyze the connectome of C. Elegans \cite{achacoso1992,watts1998} and the Macaque monkey right hemisphere~\cite{hagmann2008}. The C. Elegans network is the only complete connectome recorded consisting of the 306 neurons in the nematode worm Caenorhabditis Elegans~\cite{achacoso1992}. The network forms a directed integer weighted graph having 8,799 connections (defined by synapse or gap junctions)~\cite{watts1998}. In our analysis we treat all edges as undirected and unweighted. The Macaque monkeys connectivity between 47 regions of the right hemisphere is estimated based on diffusion spectrum imaging~\cite{hagmann2008}. We consider the undirected unweighted network where a link denotes the existence of a fiber between two regions. The network has a total of 275 undirected links. From figure~\ref{fig:CELEGANSandMonkey} it can be seen that both the IRM model as well as the unpooled and pooled hierarchical models are able to extract prominent structure defining network homogeneities. However, both the unpooled and pooled hierarchical models extract structures that are well in support of hierarchical modularity dividing the brain into parts and subparts of tightly connected groups of nodes.

\begin{figure*}
  \centering
  \includegraphics[width=\columnwidth]{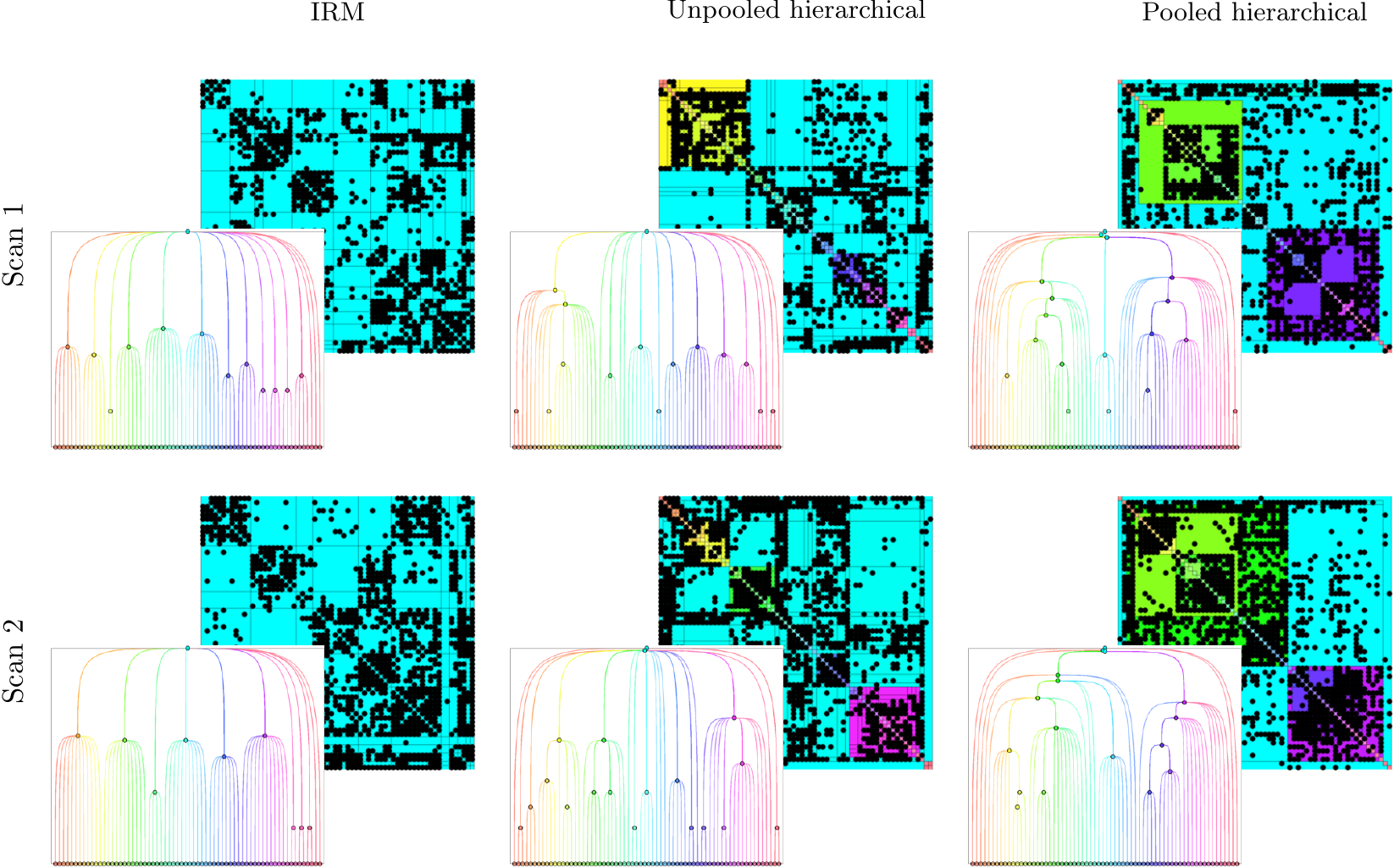}
  \caption{IRM as well as unpooled and pooled hierarchical modeling of
    the graphs derived from two separate diffusion imaging scans of
    subject 1.}
  \label{fig:subject1}
\end{figure*}

In order to quantify if hierarchical structure is supported in
structural human brain networks we consider the diffusion spectrum
imaging data\footnote{The data was downloaded from
  \url{http://connectomeviewer.org/viewer/datasets}.} described
in~\cite{hagmann2008} where we have access to multiple graphs defining
structural connectivity across five subjects. The diffusion spectrum
imaging has been used to map pathways within and across cortical
hemispheres in five human participants where the first participant has
been scanned twice. We consider the data at the resolution given by 66
anatomical gray matter regions~\cite{hagmann2008}.  We threshold the
graph such that a link exists if there is a non-zero weight in the
connectivity matrix and we model the undirected binary networks where
links indicate the existence of connectivity between two cortical
regions. The results of the modeling of the networks by the IRM model
as well as unpooled and pooled hierarchical models are given in
Figure~\ref{fig:subject1} where the analysis of the two separate scans
of subject 1 is given.
Both the IRM model as well as the pooled and unpooled
hierarchical models extract prominent network homogeneities. However,
from the results it is not clear if the hierarchical structure is significant compared to the IRM model.

To investigate this, we exploit that we have access to multiple scans,
which we consider as independent samples of the ``true'' graph of
cortical connectivity. As the first subject has been scanned twice we
use these two scans to investigating how well the models are able to
predict the graph derived from an independent scan of the same
subject. In addition, we investigate how well the network models
fitted to each subject generalize to the other subjects. We evaluate
the predictive performance in terms of the mean predictive
log-likelihood and average area under curve (AUC) of the receiver
operator characteristic (ROC) across the samples using link prediction
on the complete network of a subject (see
Table~\ref{tab:braindat}). We included the AUC score as this is a
common measure of predictive performance in networks, see also
\cite{Clauset2008,miller2009}. The IRM model performs slightly but not
significantly better than the unpooled hierarchical model on all
tasks. Both the IRM and the unpooled model substantially outperform
the pooled hierarchical model on all tasks. Thus, the data does not
support the hierarchical structure defined by the pooled hierarchy. On
the other hand, as the unpooled hierarchical model is on par with the
IRM model the inferred hierarchical structure has a relative benefits
over the IRM model in giving an interpretable representation of how
structural connectivity structure emerges at different scales.

\begin{table*}
  \centering
  \rotatebox{90}{
    \begin{tabular}{lcccccc}
        \toprule
        \multirow{2}{*}{\diagbox{Model}{Network}}
         & \multicolumn{2}{c}{Within scan} & \multicolumn{2}{c}{Within subject} & \multicolumn{2}{c}{Between subjects}                                \\
                          & $\log{L}$     & AUC              & $\log{L}$     & AUC              & $\log{L}$     & AUC              \\
\midrule
\multirow{2}{*}{IRM}      & -745.271      & 0.912719         & -1054.27      & 0.822502         & -980.321      & 0.840496         \\
                          & $\pm$ 18.7209 & $\pm$ 0.00387353 & $\pm$ 19.7743 & $\pm$ 0.0035924  & $\pm$ 10.207  & $\pm$ 0.00321595 \\
\midrule
\multirow{2}{*}{Unpooled} & -756.139      & 0.905174         & -1068.31      & 0.811378         & -1000         & 0.831984         \\
                          & $\pm$ 21.4632 & $\pm$ 0.00702501 & $\pm$ 36.3289 & $\pm$ 0.0131813  & $\pm$ 12.8714 & $\pm$ 0.0041629  \\
\midrule
\multirow{2}{*}{Pooled}   & -888.59       & 0.848447         & -1149.55      & 0.775658         & -1069.58      & 0.78661          \\
                          & $\pm$ 18.2798 & $\pm$ 0.00480982 & $\pm$ 14.0378 & $\pm$ 0.00478029 & $\pm$ 11.9004 & $\pm$ 0.00439868 \\
        \bottomrule
    \end{tabular}}
    \caption{Average log likelihood ($\log{L}$) and AUC scores within
      the same scan and the predictive log likelihood and link
      prediction AUC across the first subjects two scans (denoted
      within subject 1) and between all subjects. In parenthesis is
      given standard deviation across mean. (Within scan includes six
      samples (i.e., 5 subjects with the first subject having two
      independent scans), between scans include two samples and
      between subjects include $5\cdot4/2=10$ samples where we have
      used first scan of subject 1. }
    \label{tab:braindat}
\end{table*}

\section{Conclusion}\label{conclusion}
We have proposed to use the Gibbs fragmentation tree process as a prior over multifurcating trees in two hierarchical models for relational data. An unpooled model where individual between-group interactions where independent and a pooled model where the interaction between groups at each level of the hierarchy were assumed identical. In the analysis of synthetic networks we observed that the models well identified the structure they were designed for. In real networks we found that the hierarchical models proposed were able to model structure at multiple levels and were thereby able to model structure in clusters that by the IRM model mainly resembled noise. Furthermore the hierarchical models were able to detect structure at a resolution terminating at a more detailed level than the IRM model. Thus, the two proposed hierarchical models seem to form useful frameworks for the modeling of structure emerging at multiple levels of networks and to infer from the data the number of levels of representations needed.

The analysis of brain connectivity data in C. Elegans network as well as the right hemisphere cortical connectivity of the Macaque monkey qualitatively gave some support for the notion of hierarchical modularity as proposed in \cite{meunier2009,meunier2010,ferrarini2009,bassett2010}. However, our analysis of the human brain connectivity at the level of 66 cortical regions connectivity did not give  evidence for the presence of a hierarchical structure. For predictive modeling, the unpooled model performed on par with the IRM model while providing a hierarchical account of the structure. The lack of support for the hierarchical structure might be attributed to the low resolution of this network. With only 66 cortical regions represented finer details reflecting hierarchical structure might not be visible. Thus, in future work we will analyze structural connectivity networks at a higher resolution such as the structural connectivity of 1000 regions in the data provided by~\cite{hagmann2008}.

We believe hierarchical structure is indeed an important property of many networks including brain connectivity networks. In particular, hierarchical structure should be prominent in large scale networks where structure is likely to exist at multiple scales. Future work will focus on improving the proposed Markov chain Monte Carlo sampler for large scale inference by exploiting that the hierarchical structures inferred by the models admit sampling in parallel between the nodes belonging to separate children at given levels of the hierarchy. Furthermore, we envision the sampler at higher levels of the tree will benefit from Gibbs sampling across the multiple potential reconfigurations of the internal nodes at these higher levels.

Hierarchical modularity, i.e. systems that contain subsystems of more tightly connected nodes (that in turn may be defined by tighter connected subsystems etc.) are in the present unpooled and pooled models not explicitly accounted for. In fact, subsystems may be less densely connected than at their less detailed resolution as the density parameters are unconstrained. A benefit of keeping these parameters unconstrained is that it enable us to collapse the parameters during inference reducing the inference to sampling over tree structures. However, in future work we will aim at deriving models that explicitly accounts for hierarchical modularity using the Gibbs fragmentation tree process. Similar ideas have been proposed~\cite{morupschmidt2012} for non-hierarchical models where the within-community density is constrained be higher than between community densities while admitting analytic integration of the majority of the parameters specifying between cluster interactions. We envision this can be accomplished while preserving that the model is consistent and exchangeable by exploiting ideas from \cite{Steinhardt2012}.

The proposed framework for modeling hierarchical structure in relational data admit formal testing of hierarchical structure in networks such that the unpooled model reduce to a representation closely related to the IRM models representation when the data does not support hierarchical structure. We believe this forms a useful tool for researchers investigating and validating whether their relational systems are defined by hierarchical structures.

\bibliographystyle{plain}
\bibliography{references}

\end{document}